\documentclass[preprint,electronic]{vgtc}             

\graphicspath{{figures/}{pictures/}{images/}{./}} 

\usepackage{times}                     

\usepackage{lipsum}                    
\usepackage{tabularx}
\usepackage{tabulary}
\usepackage{capt-of}

\usepackage{multirow}

\usepackage{csquotes}

\usepackage{color, colortbl}

\usepackage{makecell}

\usepackage{booktabs,siunitx}

\usepackage{lipsum}

\usepackage{enumitem}

\usepackage{mathptmx}                  

\usepackage{subcaption}
\usepackage{xspace}
\usepackage[normalem]{ulem}
\definecolor{mred}{rgb}{.80,.12,.30}
\definecolor{MRED}{rgb}{.80,.12,.30}
\definecolor{grey}{rgb}{0.5,0.5,0.5}
\definecolor{purple}{rgb}{.75,0,.85}
\definecolor{pistachio}{rgb}{0.58, 0.77, 0.45}
\definecolor{palesilver}{rgb}{0.9, 0.9, 0.9}

\newif\ifnotes
\notestrue

\newcommand{\sys}{LinkQ\xspace}

\newcommand{\goal}[1]{\hyperref[#1]{(\textbf{G\ref*{#1}}})}

\let\origcite\cite
\renewcommand{\cite}[1]{\ifnotes\mbox{\origcite{#1}}\else \origcite{#1}\fi}

\onlineid{1193}

\vgtccategory{Research}

\vgtcinsertpkg

\ieeedoi{10.1109/VIS55277.2024.00031}

\title{\sys: An LLM-Assisted Visual Interface for Knowledge Graph Question-Answering}

\author{Harry Li\thanks{e-mail: harry.li@ll.mit.edu}\\ %
        \scriptsize MIT Lincoln Laboratory %
\and Gabriel Appleby\thanks{e-mail: gabriel.appleby@tufts.edu}\\ %
     \scriptsize Tufts University %
\and Ashley Suh\thanks{e-mail: ashley.suh@ll.mit.edu}\\ %
     \scriptsize MIT Lincoln Laboratory}

\teaser{
  \centering
  \includegraphics[width=1\linewidth]{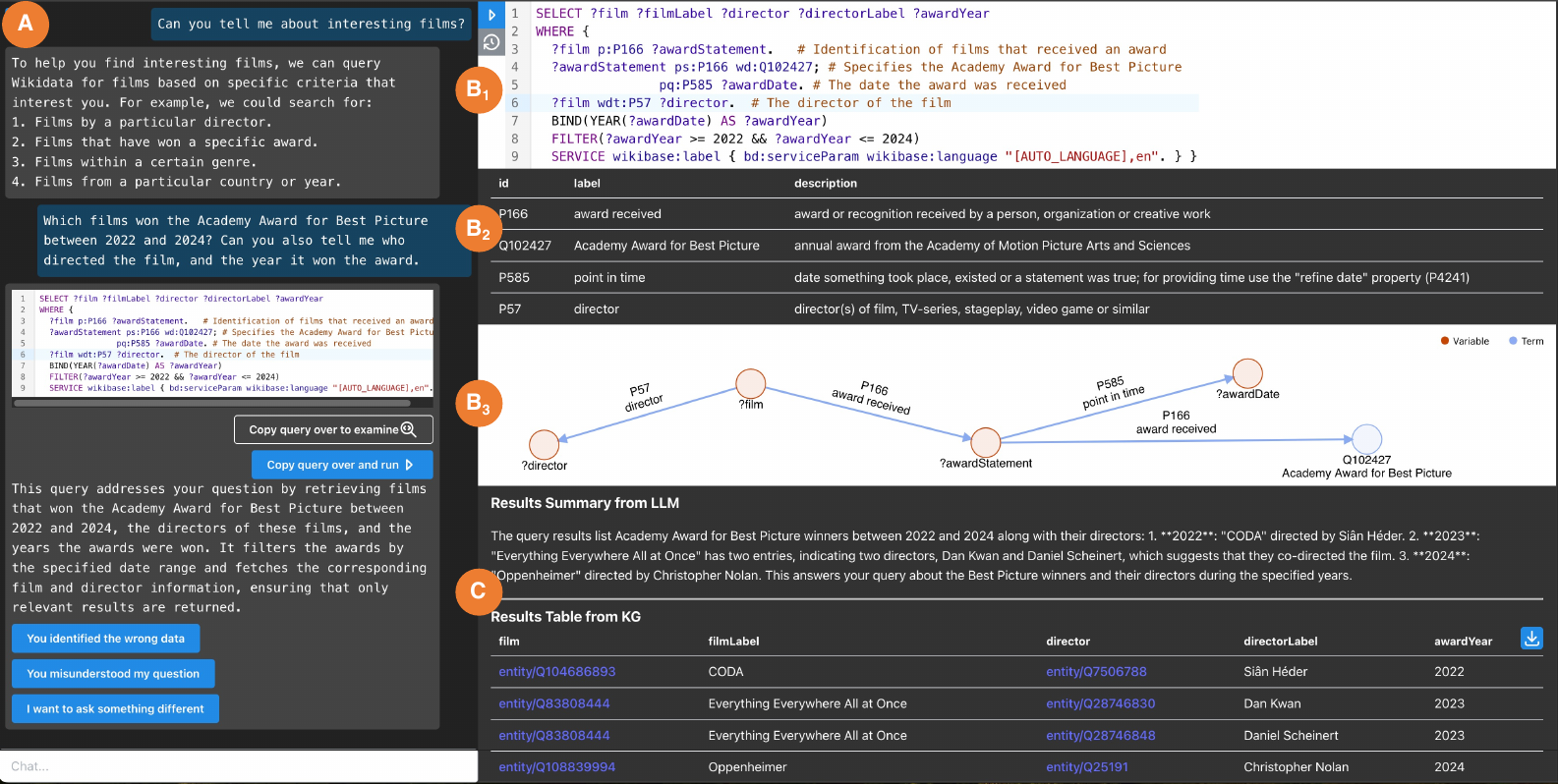}
  \caption{%
  	Exemplar workflow for \sys, a system leveraging an LLM for refining natural language questions into knowledge graph queries. The (A) \textit{Chat Panel} lets users communicate with the LLM to ask specific or open-ended questions. The \textit{Query Preview Panel} consists of three components: the (B1) \textit{Query Editor}, which supports interactive editing; the (B2) \textit{Entity-Relation Table}, which provides mapped data IDs from the KG, helping to assess the correctness of the LLM's generated query; and the (B3) \textit{Query Graph}, which visualizes the structure of the query to illustrate the underlying schema of the KG. Finally, the (C) \textit{Results Panel} provides a cleaned, exportable table as well as an LLM-generated summary based on the query results. Importantly, \sys ensures all data retrieved and summarized by the LLM comes from ground truth in the KG. 
  }
  \label{fig:teaser}
}

\abstract{%
We present \sys, a system that leverages a large language model (LLM) to facilitate knowledge graph (KG) query construction through natural language question-answering. Traditional approaches often require detailed knowledge of a graph querying language, limiting the ability for users -- even experts -- to acquire valuable insights from KGs. \sys simplifies this process by implementing a multistep protocol in which the LLM interprets a user's question, then systematically converts it into a well-formed query. 
\sys helps users iteratively refine any open-ended questions into precise ones, supporting both targeted and exploratory analysis.
Further, \sys guards against the LLM \textit{hallucinating} outputs by ensuring users' questions are only ever answered from ground truth KG data. 
We demonstrate the efficacy of \sys through a qualitative study with five KG practitioners. Our results indicate that practitioners find \sys effective for KG question-answering, and desire future LLM-assisted exploratory data analysis systems.
} 


\begin{document}

\maketitle

\section{Introduction}

Knowledge graphs (KGs) have become an industry standard for storing complex relationships in data for various tasks~\cite{abu2021domain, hogan2021knowledge}, such as recommendation generation~\cite{guo2020survey} and visualization creation~\cite{li2021kg4vis}. Despite their widespread use for managing data, a significant barrier to KG effectiveness remains the retrieval of relevant data through querying~\cite{lissandrini2020graph, lissandrini2022knowledge, li2024kgs}. This is partly because KG data retrieval can require a highly technical understanding of the KG itself or the KG's querying language~\cite{ngonga2013sorry}, which typically differs across different graph databases~\cite{khan2023querying}. When the user is a non-expert in querying, they must rely on KG query builders or traditional NL interfaces~\cite{
ell2015spartiqulation, grafkin2016sparql, yamaguchi2014intelligent, ferre2017sparklis}, which tend to limit the expressivity of the user's data analysis questions due to their rule-based syntax~\cite{karanikolas2023large}.
Consequently, valuable insights contained in KGs are difficult to attain, limiting the accessibility of KGs in practice~\cite{li2024kgs}. 

In this paper, we contribute \sys\footnote{Open-source code:   (\url{https://github.com/mit-ll/linkq})}, a system that utilizes a large language model (LLM) conversational agent to assist users in exploring and answering questions about KG data. 
Recent work has shown that LLMs are capable of interpreting analytic questions from natural language~\cite{abbasiantaeb2024let,yang2024harness} and writing code~\cite{rozière2024code}, including KG queries~\cite{rozière2024code,rangel2024sparql}. 
\sys combines these ideas, implementing SPARQL-based targeted \textit{and} exploratory question-answering with GPT-4~\cite{gpt2024stats} for the WikiData~\cite{wikidata2024stats} RDF knowledge graph~\cite{RDF}. 
When users ask open-ended questions about KG data, \sys guides the LLM in helping the user iteratively refine those questions until they can be translated into precise KG queries. 

Integrating LLMs with KGs has emerged as a promising research direction for data management applications~\cite{fernandez2023large, 2024_unifying_llms_and_kgs}. KGs are largely used to help mitigate an LLM's \textit{hallucinations}~\cite{martino2023knowledge, wen2024mindmap}, that is, the generation of false, erroneous data~\cite{rawte2023survey}. A recent survey paper details current techniques in this space~\cite{agrawal-etal-2024-knowledge}. KGs can also help an LLM access information that it was not previously trained on, as KG data can be kept consistently up-to-date without incurring the expensive retraining process for LLMs~\cite{hogan2021knowledge, wikidata2024stats}. 

\begin{figure}[!t]
\centering
  \includegraphics[width=1\columnwidth]{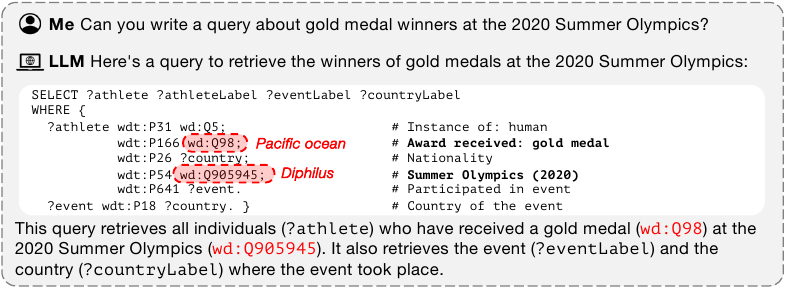} 
\caption{Example of how an LLM can \textit{hallucinate}~\cite{rawte2023survey} or misrepresent the data in a KG. Here, we ask ChatGPT~\cite{brown2020language} to write a query about gold medal winners from the 2020 Summer Olympics. In total, ChatGPT wrote three consecutive failing queries with incorrect IDs, while stating to us that it had corrected its previous mistakes.}
\label{fig:llm_hallucination}
\end{figure}

Building upon these ideas, we devise a protocol in \sys to ensure the LLM answers the user's questions by constructing and executing KG queries. Our protocol is designed to mitigate the LLM from hallucinating false data during query construction (Figure~\ref{fig:llm_hallucination}), in its outputs (Section~\ref{sec:architecture}), and allows users to retrieve recent data that may not be present in industry-standard LLMs.
For example, in Figure~\ref{fig:teaser}, the LLM (GPT-4) uses ground-truth KG data to correctly show the 2024 Best Picture winning film, which (at the time of this writing) is data it has not be trained on~\cite{gpt2024stats}.
\sys implements an interactive graph query visualization and an entity-relation table (Section~\ref{sec:interface-design}), which are populated with ground-truth data from the KG. Both of these visualizations are included to facilitate users in assessing the accuracy of the LLM's query.

To demonstrate how \sys can be used for KG query building from natural language (NL), we conducted a qualitative study with five data scientists who regularly work with KGs (Section~\ref{sec:demonstration}). In this demonstration, we walked through several use cases for the system and allowed participants to ask their own questions for exploratory analysis. Participants found high value in the workflow of \sys, particularly the LLM's explanations for its query generation process, the query graph for showcasing KG connections, and the alleviated burden of constructing KG queries altogether. To conclude, we discuss the limitations of \sys and future directions for designing LLM-driven visual analysis tools for KGs (Section~\ref{sec:future}).

\section{System Design Goals}
\label{sec:goals}
In the initial stages of designing \sys, we leveraged the learning outcomes from previous work in KG query builders~\cite{ell2015spartiqulation, grafkin2016sparql, ferre2017sparklis}, NL question-answering interfaces~\cite{srinivasan2021snowy, sultanum2023datatales, mitra2022facilitating, huang2023flownl}, as well as the feedback we received over a month-long collaboration with three data scientists who regularly use KGs in their work. 
Altogether, we arrived at the following design goals for our system:

\begin{enumerate}[topsep=2pt, partopsep=0pt,itemsep=2pt,parsep=2pt,label=\textbf{G\arabic*}]
    \item \label{goal:refine} \textbf{Support back-and-forth conversation to refine natural language questions into precise queries.} Users may have open-ended or targeted questions they want to ask about data in the KG. When the question is open-ended, an LLM should help users refine their search.

    
    \item \label{goal:no-hallucinate} \textbf{Mitigate the LLM's tendency to hallucinate false information.} Using an LLM to generate KG queries may result in false data IDs or false query results. Precautions should be made to reduce the frequency of these occurrences. 

    \item \label{goal:preview} \textbf{Preview information about the LLM's generated query to assess its accuracy.} KG queries can be computationally expensive to execute. A preview of the query, with relevant information, should be displayed to users to help assess the accuracy of both the data and the query before its execution.

    \item \label{goal:multimodal} \textbf{Provide multimodal query results from both the LLM and the KG.} Users perform question-answering with KGs for a multitude of purposes. Therefore, query results should be displayed in text form (for consumption) as well as tabular form (for further analysis). 
\end{enumerate}






\begin{figure*}[!ht]
  \centering 
    \includegraphics[width=.95\linewidth]{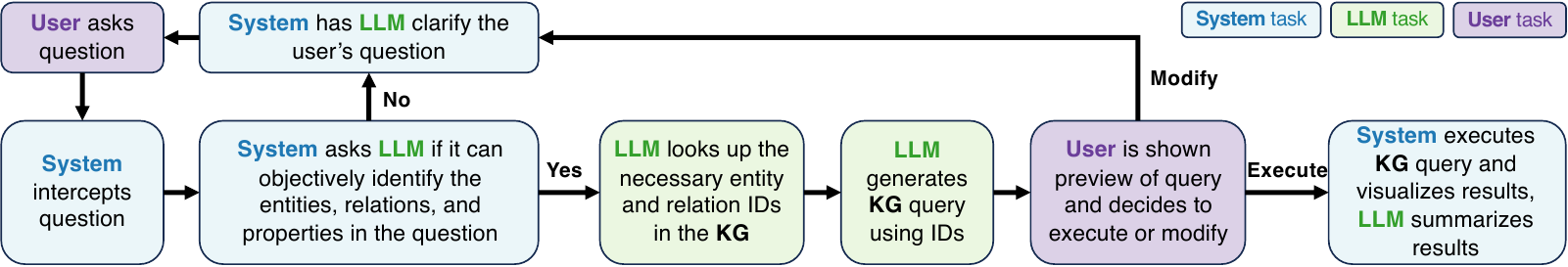}
  \caption{%
  	Illustrative state-machine of \sys's system architecture, as described in Section~\ref{sec:architecture}. In our pipeline, the \textbf{LLM}, \textbf{System}, and \textbf{User} have distinct roles, responsible for communicating with one another to complete different tasks across different states.%
  }
  \label{fig:pipeline}
\end{figure*}

\section{{\sys}}
\label{sec:architecture}
\sys follows a state-machine workflow, shown in Figure~\ref{fig:pipeline}. We provide details on the exact prompts and algorithmic workflow used for {\sys} in our supplemental material and open-source code, which are available at \url{https://github.com/mit-ll/linkq}.

An important distinction to make in \sys is the difference in responsibilities for the \textbf{LLM}, \textbf{System}, and \textbf{User}. In this section, we discuss each as separate roles: The LLM is the language model that helps process and interpret text from the User and System; the backend System relays messages, prompts, and API calls between the LLM and KG; and the User converses with the LLM without having to directly interact with the KG or System. 

\smallbreak 
\noindent 
\textbf{LLMs, Frameworks, and KGs Used:}
We tested variations of GPT~\cite{brown2020language}, Code Llama~\cite{rozière2024code}, and Mistral~\cite{jiang2023mistral} during the design of \sys, ultimately deciding on GPT-4. We found the performance of GPT-4 to be the most consistent, but any LLM (either off-the-shelf or fine-tuned) could be used in its place. Since KGs vary widely in their structure and quality~\cite{hogan2021knowledge}, we use the Wikidata KG as it is well-maintained, generalizable, and has a robust API service.

\subsection{Natural Language Question Interpretation}
To start, the System instructs the LLM that: (1) it is responsible for generating KG queries from a User's natural language questions; (2) it must help the user refine a vague, subjective, or open-ended question into a well-defined one; and (3) during refinement, it should only suggest data that it would expect to be in the KG. 

\smallbreak 
\noindent 
\textbf{Open-ended versus Targeted Question:} The User might have an open-ended question, such as ``\textit{what are some interesting things about cats?}'' When the User's question is subjective or exploratory in nature, the LLM will follow up with suggested properties or relationships from the KG that could possibly be used in place of vague qualifiers like ``\textit{interesting}.'' This back-and-forth ensures a well-formed KG query can be written (\ref{goal:refine}). 


\subsection{Finding IDs in the Knowledge Graph}
\label{sec:mapping}

LLMs are known to state plausible but erroneous information to users~\cite{rawte2023survey}. We observed this phenomenon during our initial work with various LLMs to generate KG queries -- entity and relation IDs were often incorrect. An example is illustrated in Figure~\ref{fig:llm_hallucination}.

That said, it is computationally impractical to fine-tune an LLM to memorize all possible entity, relation, and property IDs in a KG. For example, the Wikidata KG has more than a hundred million entities~\cite{wikidata2024stats}. Even if the KG were small in size, any updates to the data or IDs could result in having to retrain the LLM. 

We implemented a protocol to address this, in which the LLM iteratively asks the System to fuzzy search for entities, find relevant properties, and traverse the KG to identify all the correct IDs required to generate a SPARQL query. The System makes requests to the various KG APIs, then passes the KG graph structure and ground-truth IDs to the LLM via system prompt messages (\ref{goal:no-hallucinate}).

\smallbreak 
\noindent 
\textbf{Fuzzy Searching for Entities by Name and Context}: 
Our protocol uses a fuzzy search feature in the Wikidata API so the LLM does not have to rely on \textit{exact} names, instead identifying entity IDs based on semantic similarity. The LLM can select the entity ID that most closely matches the context of the User's question, which helps with duplicate and ambiguous entities -- a common KG data quality and querying obstacle~\cite{lissandrini2022knowledge}. For example, if the User asks a question about the \textit{box office total} for \textit{The Godfather}, the LLM should select the entity ID for \textit{The Godfather} that references the film, rather than the book series. 

\smallbreak 
\noindent 
\textbf{Given an Entity, Find its Relations and Properties}: The LLM can ask the System to get all properties and relations for a given entity ID. The System requests data from Wikidata API, then responds with the results. Importantly, this ensures that the properties actually belong to that entity in the KG. For example, if the User asks about the \textit{box office total} for \textit{The Godfather}, the LLM should look for any properties or relations belonging to \textit{The Godfather} that are named similarly to \textit{box office}. 

\smallbreak 
\noindent 
\textbf{Traverse the Graph for Multi-Hop Relationships}: To identify multi-hop relationships, the LLM can traverse the KG from one head entity to a tail entity, again by interfacing with the Wikidata API via the System. This is necessary for building successful queries, since KG data is maintained as a series of connected triplets. For example, when mapping data from the question ``\textit{Who are the Founders of Google, and what are their birthdates,}'' the LLM would have to hop from the \textit{Google} entity to its \textit{Founder} entities to identify their \textit{birthdate} properties.

\subsection{Writing, Explaining, and Executing KG Queries}
After all necessary IDs have been identified from the KG, the System instructs the LLM to generate a query to answer the original question.
The KG query is then previewed, along with an LLM-generated explanation, to the User in the interface (discussed in Section~\ref{sec:interface-design}). To produce this explanation, the System instructs the LLM to summarize why the generated query addresses the User's original question, e.g., which components of the query have been mapped from the question. 
%
%
%
\sys takes roughly 10-15 seconds to complete the protocol of converting the User's natural language question to an LLM-generated KG query and query explanation. 

\section{User Interface Design}
\label{sec:interface-design}
\sys is divided into three interface panels, shown in Figure~\ref{fig:teaser}: the Chat Panel, the Query Preview Panel, and the Query Results Panel.
\subsection{Chat Panel}
\label{sec:interface-design:chat-panel}
%
%
The Chat Panel (Figure~\ref{fig:teaser}-A) enables user communication with the LLM, maintains a record of the conversation, and optionally displays the full output generated by the LLM's message passing with the System. This panel includes features to help simplify query refinement (\ref{goal:refine}). For example, once a query is generated, two buttons are shown to the user: one to copy the query to the editor and another to copy and run the query immediately. This design prevents unintended overwriting of in-progress queries, although a query history is available for reference (see Section~\ref{sec:interface-design:query-editor}). Additional buttons are provided to insert convenient prompts into the chat, helping users correct or redirect the LLM as needed.
\subsection{Query Preview Panel}
\label{sec:interface-design:query-preview-panel}
The Query Preview Panel comprises the Query Editor, Entity-Relation Table, and the Query Graph visualization. These three pieces work in tandem to preview the potentially computationally expensive query before it is run, and ensure the accuracy of Wikidata entity identifiers (\ref{goal:preview}).
\smallbreak 
\noindent
\textbf{Query Editor}
\label{sec:interface-design:query-editor}
%
The Query Editor (Figure~\ref{fig:teaser}-B1) is a standard SPARQL code interface with keyword highlighting, helping users assess the syntax of the query (\ref{goal:preview}).
The editor also features a query history, allowing users to quickly retrieve past queries.
\smallbreak 
\noindent
\textbf{Entity-Relation Table}
\label{sec:interface-design:entity-relation-table}
The Entity-Relation Table (Figure~\ref{fig:teaser}-B2) provides contextual information (e.g., a human-readable label and short description) about the entity and property IDs found in the query (\ref{goal:preview}). 
This table is intended to help users assess whether the LLM has identified the correct IDs for the generated query.
%
\smallbreak 
\noindent
\textbf{Query Graph}
\label{sec:interface-design:query-graph}
The Query Graph (Figure~\ref{fig:teaser}-B3) shows a visual preview (\ref{goal:preview}) of a syntactically valid query.
This graphic allows the user to understand the relationships within their queries better, regardless of whether the queries are constructed by the LLM or themselves. 
%
Specifically, the query is parsed for basic graph patterns (BGP), and the triples contained within the BGPs are plotted within the graph. All items and data values are represented as nodes, and all properties are represented as edges. The nodes and edges are labeled by their corresponding text from the query, as well as labels retrieved from the KG. Known Wikidata entities are colored blue, and unresolved variables are colored orange.
%
\subsection{Query Results Panel}

The Query Result Panel (Figure~\ref{fig:teaser}-C) provides the results of the query in formats best for both consumption and analysis (\ref{goal:multimodal}).

\smallbreak 
\noindent
\textbf{Exportable Results Table}
%
Query results are automatically parsed from JSON to a cleaned CSV format, removing unnecessary metadata retrieved from the KG, then visualized as a table. Users can choose to export the results for downstream analysis (\ref{goal:multimodal}).

\smallbreak 
\noindent
\textbf{LLM Results Summary}
The LLM also provides a 
succinct paragraph to summarize the query results for the user (\ref{goal:multimodal}). When the query results are empty, \sys instructs the LLM to provide a best guess for what could have gone wrong with the query. 

\begin{table}[]
\renewcommand{\arraystretch}{1.1}
\small 
\begin{tabular}{p{0.15\linewidth} p{0.75\linewidth}}
\toprule 
\textbf{Use Case} & \textbf{Natural Language Question to Translate} \\
\midrule
Films & (Q1) Which directors won the \textit{Academy Award for Best Director} between 2014 and 2024, and for which films?  \\
 & (Q2) Which film genres most commonly win the \textit{Academy Award for Best Picture}?  \\
\hline 
Cybersecurity & (Q3) What are some different types of cyberattacks? \\
 & (Q4) What are examples of cyberattacks that have occurred in history? \\
\hline 
Geography & (Q5) What is the official language for each country in Europe? \\
 & (Q6) What are the top 10 tallest mountains in the world, and what country do they belong to? \\
\bottomrule
\end{tabular}
\caption{Questions we walked through with participants during the demonstration of \sys. More details in Section~\ref{sec:demonstration}.} 
\label{tab:demo-questions}
\end{table}

\section{Demonstration}
\label{sec:demonstration}

To gather feedback on our system, we demonstrated its question-answering capabilities with a team of five data scientists (two we previously collaborated with, see Section~\ref{sec:goals}) who work with KGs.
Our goal was to determine whether \sys satisfies its design goals, and to identify features that could be improved in the future.

We demoed three use cases with participants. All questions from the demonstration are provided in Table~\ref{tab:demo-questions}. Participants were also asked to input their own questions to \sys. 
Our supplemental provides complete details on our protocol, our participants' experience levels, and the correct answers to each of the questions asked.

\subsection{Question-Answering Results}

Participants observed two inaccuracies. First, the LLM mistakenly used the ID for `\textit{Best Director}' instead of for `\textit{Best Picture,}' which participants identified using the Entity-Relation table (\ref{goal:preview}). This mistake was corrected after participants iterated with the LLM. In Q6, the query was syntactically correct but incorrectly resulted in U.S. mountains, likely due to a unit conversion error between feet and meters, as cautioned by the LLM's Summary Panel.

During the free-form Q\&A, participants intentionally asked questions they thought \sys would fail in translating. To their surprise, in 2/3 cases our system successfully resulted in syntactically correct SPARQL queries and correct IDs. 
Failures were due to missing data in Wikidata or incorrect unit specifications in queries.

Overall, participants praised \sys's ability to preserve their semantic intent during question-answering (\ref{goal:refine}), and told us that -- even when the LLM's query failed, our Query Graph gave them the insight to update the query correctly themselves. 

\subsection{Participants' Feedback on \sys}
After the demonstration, we performed a semi-structured interview with participants to discuss the system features that seemed most helpful in answering questions, as well as which features could be added to improve usability. We highlight interesting observations and takeaways from participants to help inform this design space.

\smallskip
\noindent 
\textbf{With enough guidance, LLMs can alleviate the burden of manual data retrieval:} Without hesitation, the immediate feedback we received was, ``\textit{What I like about this is that I don't have to write SPARQL queries anymore!}'' Participants suggested it would be extremely beneficial to data scientists if our system could also retrieve data from multiple enterprise Wikis and databases simultaneously.  

\smallskip
\noindent 
\textbf{The KG-generated Entity-Relation table and Graph Query increase trust in the LLM's outputs:} 
All participants told us that knowing the table and query graph were extracted from the KG gave them confidence in the LLM's queries (\ref{goal:no-hallucinate}). 
The two participants who work most closely with KGs praised the graph query visualization: ``\textit{%
You see edge connections that you wouldn't otherwise see from just the entity-relation table.}'' Participants less familiar with KGs preferred the table, particularly the data descriptions. 

\smallskip
\noindent 
\textbf{The LLM Summary Panel can inform what went wrong with the query, or the query results:} Participants told us, ``\textit{The LLM summary is helpful and informative\ldots it gives us information on what could've gone wrong in the query. It captured the intention of my query, which was unexpected.}'' 
This feedback suggests that there are benefits to using LLMs over traditional NLP approaches for query building. Of course, the quality of an LLM's insights depends on its own knowledge of the data domain. 

\smallskip
\noindent 
\textbf{Question-answering with ground truth data is seen as a novel task for LLMs:}
One participant told us, ``\textit{This is one of the cooler applications I've seen of LLMs. It's retrieving actual, ground truth information. Usually LLMs are just text generators.}'' Participants stressed again that research should continue to examine how LLMs can be used to improve search and data retrieval tasks for data scientists, as manual keyword search is both tedious and flawed. 

\smallskip
\noindent 
\textbf{Make visual distinctions between LLM output and KG output:} One participant told us that he sometimes forgot which outputs came from the LLM versus the KG, and suggested that a visual distinction (e.g., different colored background panels) should be added. All participants agreed that both the LLM summary and KG results table seem equally important, just categorically different. 


\smallskip
\noindent 
\textbf{Additional visualizations seemed unnecessary:} We asked participants if they wanted or expected to see additional visualizations after receiving query results. Surprisingly, not a single participant said yes, and instead told us the LLM's text summary and KG results table were ideal (\ref{goal:multimodal}). Two participants told us that it depends on the data and task, e.g., an organizational data domain could benefit from hierarchical-based charts. 

\section{Discussion and Future Work}
\label{sec:future}
The positive feedback we received from participants suggests that systems like \sys can be successfully used for question-answering over knowledge graphs. That said, it is important to note that we observed a great deal of necessary hand-holding in implementing an LLM for successful KG query writing. 
Of course, this makes sense, as off-the-shelf LLMs are typically performant on text-based tasks like summarization~\cite{yang2024harness}. Future work can investigate optimizing our approach (Section~\ref{sec:architecture}) with this in mind.




\smallskip
\noindent 
\textbf{LinkQ's Protocol versus RAG:}
Retrieval augmented generation (RAG)~\cite{rag} approaches use text embeddings to compare the user's question with a document knowledge base, and inject the most relevant documents into a prompt to improve the LLM's response. In contrast, LinkQ instructs the LLM to iteratively explore the KG data (with the help of the System) to find the IDs it needs to generate a query.
In the future, we would like to quantitatively evaluate the effectiveness of \sys against traditional RAG and NLP approaches. We can quantitatively compare the goodness of generated queries, as well as users' feedback on how well the LLM captures their semantic intent during question-answering. Altogether, these directions can help us better understand how systems like \sys can aid data scientists in conducting visual data analysis with LLMs.
\section{Conclusion}
We presented \sys, a system that implements an LLM for KG question-answering without the expertise of a KG querying language. \sys supports users in asking natural language questions, either open-ended or targeted, which are iteratively refined with an LLM until they can be translated into well-formed KG queries. \sys implements a scalable protocol to reduce the frequency of LLMs hallucinating data from the KG, requiring no fine-tuning of the LLM. Our interface provides two types of query visualizations, an entity-relation table and query graph structure, to help users both ascertain the accuracy of the LLM's generated query, and to inform them of how a query could be incorrect or improved upon. From a demonstration with five data scientists, we found that \sys exceeded expectations on converting NL questions to SPARQL queries, which participants stressed is a huge improvement to the burden of manually writing and assessing KG queries.

\acknowledgments{
We thank our collaborators at MIT Lincoln Laboratory for their helpful feedback on LinkQ, as well as the reviewers for their input in improving the quality of our paper.

\smallbreak 
\noindent 
DISTRIBUTION STATEMENT A. Approved for public release. Distribution is unlimited.
This material is based upon work supported by the Combatant Commands under Air Force Contract No. FA8702-15-D-0001. Any opinions, findings, conclusions or recommendations expressed in this material are those of the author(s) and do not necessarily reflect the views of the Combatant Commands. © 2024 Massachusetts Institute of Technology. Delivered to the U.S. Government with Unlimited Rights, as defined in DFARS Part 252.227-7013 or 7014 (Feb 2014). Notwithstanding any copyright notice, U.S. Government rights in this work are defined by DFARS 252.227-7013 or DFARS 252.227-7014 as detailed above. Use of this work other than as specifically authorized by the U.S. Government may violate any copyrights that exist in this work.
}

\bibliographystyle{abbrv-doi}

\bibliography{template}
\end{document}


\maketitle

\section{Demonstration}
\label{sec:demonstration}

\section{Workflow}
\label{sec:workflow}

\section{Large Prompts}
\label{sec:large-prompts}
\subsection{Initial System Message}
\label{sec:large-prompts:initial-system}
The initial system message primes the large language model (LLM) for its duties within the system.
\begin{verbquote}
You are a helpful chat assistant. This system will give you access to data in the WikiData Knowledge Graph, that contains encyclopedic data similar to Wikipedia, but in knowledge graph format using the RDF framework. 

If users ask questions that can be answered via WikiData, your job is not to directly answer their questions, but instead to help them write a SPARQL query to find that data. You can ask the user to clarify their questions if the questions are vague, open-ended, or subjective in nature. 

If you ever need to suggest data to the user, you should only provide recommendations that are directly accessible from Wikidata. Do not ask the user if they would like to proceed with generating the corresponding query unless absolutely necessary.

When you are ready to start building a query, respond with 'BUILD QUERY'. The system will walk you through a guided workflow to get the necessary entity and property IDs from WikiData.

Current date: {date}.
\end{verbquote}
\subsection{Initial ID Identification Message}
\label{sec:large-prompts:initial-id-identification}
This message asks the LLM to construct a list of items it would like to search Wikidata for, asking it to annotate it responses by whether or not it would like to look for an entity, or the property of an entity.
\begin{verbquote}
Your goal is to find the necessary entity and property IDs to construct a SPARQL query that answers the user's question. Do not respond with a trailing period. Do not assume you already know the correct entity and property IDs; you should search for them. Respond in one of these ways:
- To fuzzy search for an entity, start the response with 'ENTITY SEARCH:', followed by an entity name you want to search for. The system will respond with possible entity resolutions in Wikidata. 
- To get all the properties for an entity, start the response with 'PROPERTIES SEARCH:', followed by the ID of the entity. The user will respond with all the properties associated with that entity.
- To find what entities are connected to the original entity via a property, start the response with 'ENTITY PROPERTY SEARCH:', followed by the entity ID then the property ID. Ex: 'ENTITY PROPERTY SEARCH: Q123 P456'
- Respond with 'STOP' if and only if you have searched for and succesfully identified all necessary IDs from Wikidata to construct the query.
\end{verbquote}
\subsection{Query Message}
\label{sec:large-prompts:query-message}
This prompt tells the LLM to construct a SPARQL query, giving it relevant background about Wikidata, and also performing few-shot learning.
\begin{verbquote}
You are an expert at generating SPARQL queries for the Wikidata Knowledge Graph from natural language. 
Entity IDs are prepended with 'wd' and property IDs are prepended with 'wdt'. 
Your task is to convert the natural language instruction into a SPARQL query.
The following are four examples in which I am showcasing a natural language instruction (NLI) and the converted SPARQL Query. 
  NLI: Who are creators of Apple and what are their birthdates?
  SPARQL Query:
    SELECT ?founder ?founderLabel ?birthdate
      WHERE {
        wd:Q312 wdt:P112 ?founder.   # Q312 represents Apple and P112 represents founder
        ?founder wdt:P569 ?birthdate. # P569 represents date of birth
        
        SERVICE wikibase:label { bd:serviceParam wikibase:language "[AUTO_LANGUAGE],en". }
    }
  NLI: Who are the current heads of state for all countries in the world? 
  SPARQL Query: 
    SELECT ?country ?countryLabel ?headOfState ?headOfStateLabel 
      WHERE { 
        ?country wdt:P31 wd:Q6256;     # Instance of: country 
        p:P35 ?statement.    # has head of government statement 
        ?statement ps:P35 ?headOfState;   # head of government property 
        pq:P580 ?startDate.   # start date of the term 
        FILTER NOT EXISTS { ?statement pq:P582 ?endDate }  # Ensure current head of state 
        SERVICE wikibase:label { bd:serviceParam wikibase:language "[AUTO_LANGUAGE],en". } 
      } 
      ORDER BY ?countryLabel 
  NLI: What are the top five tallest mountains in the world and their respective heights? 
  SPARQL Query: 
    SELECT ?mountain ?mountainLabel ?height 
      WHERE { 
        ?mountain wdt:P31 wd:Q8502;         # Instance of: mountain 
        wdt:P2044 ?height.       # Height property 
        FILTER (?height >= 8000)           # Minimum height of 8000 meters 
        SERVICE wikibase:label { bd:serviceParam wikibase:language "[AUTO_LANGUAGE],en". } 
      } 
    ORDER BY DESC(?height) 
    LIMIT 5 
  NLI: Which symphonies were composed by Ludwig van Beethoven? 
  SPARQL Query: 
    SELECT ?composition (SAMPLE(?compositionLabel) as ?compositionLabel) 
      WHERE { 
        ?composition wdt:P31 wd:Q105543609;         # Instance of: Beethoven's symphonies 
        wdt:P86 wd:Q255;               # Composer: Ludwig van Beethoven 
        rdfs:label ?compositionLabel. 
        FILTER(CONTAINS(LCASE(?compositionLabel), "symphony")) 
        SERVICE wikibase:label { bd:serviceParam wikibase:language "[AUTO_LANGUAGE],en". } 
      }  
    GROUP BY ?composition 
  Start the SPARQL query with \`\`\`sparql and end the query with \`\`\`. After you generate a SPARQL query, you briefly explain, as concisely as possible, to the user why the query addresses their original question. Keep your explanation as short as possible and only further explain when asked.

  Now construct a query that answers the user's question: {text}
\end{verbquote}